\documentclass[journal]{IEEEtran}
\ifCLASSINFOpdf
\else
\fi
\usepackage{amsmath,amssymb,amsfonts}
\usepackage{algorithmic}
\usepackage{graphicx}
\usepackage{textcomp}
\usepackage{xcolor}
\usepackage{bm}
\usepackage{bbm}
\usepackage{todonotes} 
\usepackage{multirow}
\usepackage{booktabs}
\usepackage{algorithm}
\usepackage{microtype}
\usepackage{eqparbox}
\usepackage{url}
\usepackage{subcaption}
\usepackage{relsize}
\usepackage{hyperref}
\usepackage{textcomp}
\usepackage{array}
\newcolumntype{P}[1]{>{\centering\arraybackslash}p{#1}}

\usepackage[utf8]{inputenc}
\DeclareUnicodeCharacter{0308}{}
\DeclareUnicodeCharacter{0301}{}

\usepackage[backend=biber,style=ieee,sorting=none, maxbibnames=10]{biblatex}
\usepackage{microtype}

\renewcommand{\algorithmiccomment}[1]{\bgroup\hfill$\triangleright$~#1\egroup}

\newcommand{\argmax}[1]{\underset{#1}{\operatorname{arg}\,\operatorname{max}}\;}

\begin{document}


%
\title{Student-Initiated Action Advising\\ via Advice Novelty}
%
%
%

\author{Ercument~Ilhan,
        Jeremy~Gow
        and~Diego~Perez-Liebana
        
        
}

\maketitle

\begin{abstract}
Action advising is a budget-constrained knowledge exchange mechanism between teacher-student peers that can help tackle exploration and sample inefficiency problems in deep reinforcement learning (RL).
Most recently, student-initiated techniques that utilise state novelty and uncertainty estimations have obtained promising results. 
However, the approaches built on these estimations have some potential weaknesses.
First, they assume that the convergence of the student's RL model implies less need for advice.
This can be misleading in scenarios with teacher absence early on where the student is likely to learn suboptimally by itself; yet also ignore the teacher's assistance later.
Secondly, the delays between encountering states and having them to take effect in the RL model updates in presence of the experience replay dynamics cause a feedback lag in what the student actually needs advice for.
We propose a student-initiated algorithm that alleviates these by employing Random Network Distillation (RND) to measure the novelty of a piece of advice.
Furthermore, we perform RND updates only for the advised states to ensure that the student's own learning does not impair its ability to leverage the teacher.
Experiments in GridWorld and MinAtar show that our approach performs on par with the state-of-the-art and demonstrates significant advantages in the scenarios where the existing methods are prone to fail.
\end{abstract}



%
\IEEEpeerreviewmaketitle

\section{Introduction} \label{sec:introduction}

\IEEEPARstart{D}{eep} reinforcement learning (RL) has achieved outstanding feats in a wide range of domains including board games \cite{DBLP:journals/corr/abs-1712-01815}, complex video games \cite{alphastarblog}\cite{DBLP:journals/corr/abs-1912-06680} and robotics \cite{DBLP:journals/jmlr/LevineFDA16} in the recent years.
In addition to its remarkable performance, deep RL's end-to-end structure and general applicability make it a desirable decision-making tool for many real-world problems.
Nevertheless, such advantages come at the cost of long training times and large numbers of samples acquired through environment interactions, which are well-known drawbacks of deep RL \cite{DBLP:journals/corr/abs-1708-05866}. 
There is currently a significant amount of research effort focused on improving exploration techniques to make agents more proficient at discovering the environment mechanics and collecting samples that will yield high-reward strategies \cite{DBLP:journals/corr/abs-1908-02388}.
Alternatively, in some situations, agents may also have access to legacy knowledge of some peers, be it humans or other agents.
This presents a whole new set of opportunities to tackle the exploration challenges by leveraging these.

Peer-to-peer knowledge transfer techniques have been investigated in various forms in RL to accelerate the learning processes \cite{DBLP:journals/aamas/SilvaWCS20}.
The applications within deep RL can be examined in two main groups as follows: learning from demonstration (LfD) \cite{DBLP:conf/nips/Schaal96} and action advising \cite{DBLP:conf/atal/TorreyT13}.
In LfD, the agent has access to some previously collected data by one \cite{DBLP:conf/aaai/HesterVPLSPHQSO18} or more \cite{DBLP:journals/corr/abs-1910-12154} peers to be leveraged in a pre-training stage or during the actual RL stage to improve its learning performance.
Action advising, however, does not deal with the previously collected data, and instead, is concerned with devising techniques for the situations where the agent has access to a competent peer that can provide advice in the form of actions upon request.
This is similar to the active learning paradigm in supervised learning \cite{settles2009active}.
As the main focus of this study, action advising concept is especially important in the scenarios where it is costly or not possible at all to generate useful expert data in advance, e.g., when the relevant task specifications are unknown a priori. 

The initial problem setup of the action advising framework \cite{DBLP:conf/atal/TorreyT13} assumed the learning agent, namely, the student, to be monitored constantly by a teacher who is responsible to manage the distribution of these advice. 
However, this is impractical because of possible limitations in communication and the teacher's attention span.
Moreover, the teacher has no access to the student's internal model, which renders its advising decisions to be determined solely by the teacher's initiative.
Therefore, this framework has also been extended into different versions such as student-initiated and jointly-initiated, letting the student take an active role in initiating these interactions, as covered in Section~\ref{sec:related_work}.
Despite its potential advantages, it is far from trivial to devise efficient student-initiated action advising strategies in deep RL, as the limited number of the studies may suggest.

In deep RL, assigning credit to a transition for its long-term contribution to the learning progress is a very challenging task.
Accordingly, it is also difficult to determine the actual importance of a state to obtain expert advice for.
Therefore, the student-initiated approaches in deep RL currently follow heuristics as proxies of potential learning contribution, such as novelty \cite{DBLP:conf/atal/SilvaGC17}\cite{DBLP:conf/cig/IlhanGP19} and uncertainty \cite{DBLP:conf/aaai/SilvaHKT20a}\cite{DBLP:journals/corr/abs-1812-02632} estimations.
Despite demonstrating promising performances by simplifying challenging characteristics of deep RL, these techniques have several drawbacks.
First, these estimations become smaller in correlation with the student's RL model convergence, which means that the student becomes less likely to request advice for the states it has already encountered.
Even though this behaviour is intended, it may be problematic in some scenarios.
For instance, the teacher may not be accessible early in the training due to communication related issues (that can occur in real-world domains) or because it joins the training at some later time just like in a multi-agent system.  
Moreover, if the teacher also has a non-stationary policy, it may refuse to give any advice in some stages of the learning session, which can be considered as absence.
In this case, despite the learning progression of the student's model, it will lack an adequate amount of expert contribution and will face the risk of remaining suboptimal.
Consequently, even when the teacher becomes available at a later timestep, it will be ignored by the student due to the aforementioned property of these approaches; hence, the student will fail to benefit from the advice to improve its policy further.
Apart from this weakness, if the student's learning algorithm incorporates an experience replay mechanism as in off-policy RL, e.g., Deep Q-Network (DQN) \cite{DBLP:journals/corr/MnihKSGAWR13}, it takes some time between encountering a state and involving it in the model update steps for a sufficient number of times to have it influence the state novelty and uncertainty estimations.
As result, these delays may mislead the student about what advice it actually needs to collect, especially since the contents of its replay memory is disregarded by these action advising methods.

In this study, we first highlight and demonstrate some drawbacks of the existing student-initiated action advising methods.
Then, we propose a state novelty based method involving Random Network Distillation (RND) \cite{DBLP:journals/corr/abs-1810-12894}, which, unlike the previous studies, is exclusively updated for the states that are advised (hence the name \emph{advice novelty}).
In other words, the mechanism that drives the decisions of the student's advice queries gets to be affected by only these states.
Thus, it is ensured that the student will not ignore the teacher regardless of its own state of learning unless it has already utilised the teacher's advice.

Finally, RND updates are performed with single samples in an on-line fashion instead of with batches off a replay memory.
This prevents the RND model from converging to a global optima, which gives the earlier states a chance to be asked for advice again periodically.
This is similar to the idea of keeping expert transitions in the replay memory over the course of learning to be benefited from occasionally, as in \cite{DBLP:conf/aaai/HesterVPLSPHQSO18}.
Because, no matter how many times the expert actions are executed, there is no guarantee for the expert behaviour demonstrated in them to be completely distilled into the student’s policy.
This way of RND learning minimises the possibility of the advice to be discarded forever without providing enough contribution.
Additionally, this can be advantageous if the teacher happens to have a non-stationary policy that generates advice actions.
In this case, getting advised by the teacher in different stages of learning will yield different outcomes, which is essential to learn from everything the teacher has to offer.

This paper is structured as follows: 
Section~\ref{sec:related_work} provides a review of the related work. 
Then, the relevant background information are provided in Section~\ref{sec:background}.
The game environments used in this study as presented in Section~\ref{sec:game_envs}.
In Section~\ref{sec:proposed_method}, we describe our proposed method in detail.
The experiment setup is explained in Section~\ref{sec:experimental_setup}.
Afterwards, the results are presented and discussed in Section~\ref{sec:results}.
Finally, the study is concluded with some final remarks in \ref{sec:conclusions}.

\section{Related Work} \label{sec:related_work}

Action advising was formalised in \cite{DBLP:conf/atal/TorreyT13} for the first time as teacher-student framework. 
According to this, a student agent is observed constantly by a teacher agent to be given action advice in order to make it achieve the best possible learning performance.
However, the number of advice requests is limited with a budget constraint due to practical concerns.
The problem of distributing this budget is addressed with several heuristics they proposed, such as early advising and importance advising.
Following the proposed ideas, several different approaches in this line of work have emerged.
In \cite{zimmer2014teacher}, the teacher-initiated action advising problem was treated as a meta-level RL problem.
By training the student agent over multiple learning sessions, the teacher tries to learn the most appropriate times to advise by using the student's learning state as reward feedback.
Another study \cite{DBLP:conf/ijcai/AmirKKG16} extended the initially proposed heuristic-based idea to student-initiated and jointly-initiated forms by devising several new heuristics.
In \cite{DBLP:conf/ijcai/ZhanBT16}, the case of having multiple teachers is investigated, as well as dealing with situations where the teacher's expertise may be suboptimal.
Using RL to achieve optimal advising was also studied in \cite{DBLP:journals/make/FachantidisTV19} similarly to \cite{zimmer2014teacher}. 
As well as learning \emph{when} to advise, the teachers are also made to learn \emph{what} to advise in this work.
\cite{DBLP:conf/atal/SilvaGC17} applied heuristic-based action advising to the domain of multi-agent RL with cooperative agents.
The agents are made to determine their expertise by counting state visits to take on student or teacher roles dynamically to exchange knowledge with their peers to improve team-wide learning.

Deep RL is a relatively new application domain for action advising methods.
\cite{DBLP:conf/aaai/OmidshafieiKLTR19} is one of the first studies in this particular subject.
Even though it was based on the agents with classical RL methods, the teaching via action advising problem is tackled as a meta-level deep RL problem to determine the actions to be advised as well as their temporal distribution.
Similarly, \cite{DBLP:journals/corr/abs-1903-03216} further developed this idea to also be able to handle agents that employ deep RL. 
Despite being effective in forming teacher-student interactions bidirectionally without any fixed roles, these approaches share the drawbacks of requiring the agents to be trained over multiple, centralised learning sessions due to the meta-RL mechanisms.
This may also cause the agents to end up being tuned for each other and face difficulties when paired with different peers.

Another line of work in deep RL involves applying action advising methods by following heuristics without any pre-training, which forms the specific group of approaches our study belongs to.
In \cite{DBLP:journals/corr/abs-1812-02632}, the LfD paradigm was combined with active learning by making the student agent query for demonstrations itself in a very similar way to action advising.
To do so, they make use of some specialised network architectures that make it possible for the student agent to measure its state uncertainty, and be able to use this estimation to determine the most appropriate times and states to request advice for.
\cite{DBLP:conf/cig/IlhanGP19} further developed the idea of \cite{DBLP:conf/atal/SilvaGC17} to make the agents initiate teacher-student interactions in multi-agent deep RL.
Instead of state counting, RND is employed to compute the novelty of the states to be compatible with complex state spaces that require non-linear function approximation.
More recently, \cite{DBLP:conf/aaai/SilvaHKT20a} utilised uncertainty measurements obtained via multiple neural network heads in student-initiated action advising to time the advice requests similarly to \cite{DBLP:journals/corr/abs-1812-02632}; however, they don't employ a special loss function as in LfD.
In parallel to these studies, \cite{aawithimitation} introduced an approach to imitate and reuse the teacher advice in order to prevent spending budget on previously queried states, which can be combined with other action advising methods to make them more practically feasible.

In addition to the aforementioned drawbacks, none of these heuristic-based methods for deep RL is designed to be able to handle the extended absence of the teacher.
As we highlight in Section~\ref{sec:introduction}, this is especially crucial in real-world applications.
Our study aims to provide a solution that is suitable for such scenarios, as well as alleviating the present individual drawbacks of requiring uncertainty models \cite{DBLP:conf/aaai/SilvaHKT20a}\cite{ DBLP:journals/corr/abs-1812-02632}, being susceptible to experience replay induced latency in estimations \cite{DBLP:conf/cig/IlhanGP19, DBLP:conf/aaai/SilvaHKT20a, DBLP:journals/corr/abs-1812-02632} which are detailed in Section~\ref{sec:proposed_method}.

\section{Background} \label{sec:background}

\subsection{Reinforcement Learning and Deep Q-Networks} \label{subsec:dqn}

Reinforcement learning (RL) is a prominent framework for solving sequential decision-making tasks that involve an agent acting in an environment that is formalised by a \emph{Markov decision process} (MDP).
In an MDP, an environment is defined by a tuple $\langle S,A,R,T\rangle$, where $S$ is finite set of states, $A$ is finite set of actions, $R\colon S \times A \times S \rightarrow \mathbb{R} $ is reward function and $T\colon S \times A \rightarrow \Delta(S) $ is transition function.
Acting in an MDP happens according to the agent's policy $\pi\colon S \rightarrow A$ which maps states to actions.
At each timestep $t$, the agent observes the state $s_t$ and executes the action $a_t$ in order to receive the reward $r_t$ and advance to the next state $s_{t+1}$.
The objective of RL algorithms is to obtain the $\pi^*$ that maximises expected sum of discounted rewards $\sum_{k=0}^{T} \gamma^{k} r_{t + k}$ obtained from timestep $t$ over a horizon $T$ ($\gamma \in [0,1]$ is the discount factor).

Deep Q-Network (DQN) \cite{DBLP:journals/corr/MnihKSGAWR13} is a scaled-up version of the well known fundamental RL technique Q-learning with non-linear function approximation suitable for complex domains, which attempts to obtain an optimal policy by learning state-action values $Q(s,a)$ in an end-to-end fashion.
To do so, it utilises a neural network with weights $\theta$ to map an input state $s$ to $Q(s,a)$ by minimising the loss term $(r_{k+1} + \gamma \max_{a'} Q_{\bar{\theta}}(s_{k+1}, a') -  Q_{\theta}(s_{k}, a))^2$ via stochastic gradient descent over randomly sampled transitions from timestep $k$.
Furthermore, DQN also utilises certain mechanisms to achieve functionality.
One of these is keeping the network weights $\theta$ in a separate copy network that is updated periodically and is used as a reference point in updates to stabilise learning.
Another critical component of DQN is called experience replay.
It stands for the process of saving the encountered transitions in a replay memory to be used in the model updates later on.
Not only such off-policy updates improve sample efficiency, but also this mechanism helps to break the non-i.i.d. property of the samples to make convergence easier.

Following its breakthrough, DQN is enhanced with additional techniques over years that are combined under the name of Rainbow DQN \cite{DBLP:conf/aaai/HesselMHSODHPAS18}.
In our study, we employ double Q-learning \cite{DBLP:conf/aaai/HasseltGS16}, dueling networks \cite{DBLP:conf/icml/WangSHHLF16} and noisy networks \cite{DBLP:conf/iclr/FortunatoAPMHOG18} among these, which we believe are the most essential ones.
Despite being a substantial enhancement, Prioritised Experience Replay \cite{DBLP:journals/corr/SchaulQAS15} is not utilised in our work, since it may amplify the effects of having more important samples in the replay memory, and consequently make the comparison between different action advising methods less impartial.
Hence, uniformly random sampling is chosen as the default experience replay approach.

\subsection{Noisy Networks} \label{subsec:noisy_networks} 

Noisy Networks (NoisyNets) \cite{DBLP:conf/iclr/FortunatoAPMHOG18} are an eminent exploration technique for deep RL algorithms, which is also employed in Rainbow DQN \cite{DBLP:conf/aaai/HesselMHSODHPAS18} as the default exploration strategy.
In principle, NoisyNets is a modified linear layer with noise perturbations as follows:

\begin{equation}\label{noisy_layer}
    y = (\mu_w + \sigma_w \odot \epsilon_w)x + \mu_b + \sigma_b \odot \epsilon_b
\end{equation}

where the input is $x \in \mathbb{R}^{p}$, the output is $y \in \mathbb{R}^q$, the learnable parameters are $\mu_w \in \mathbb{R}^{q \times p}$, $\sigma_w \in \mathbb{R}^{q \times p}$, $x \in \mathbb{R}^{p}$, $\mu_b \in \mathbb{R}^{q}$, $\sigma_b \in \mathbb{R}^{q}$, and the noise parameters are $\epsilon_w \in \mathbb{R}^{q \times p}$, $\epsilon_b \in \mathbb{R}^{q}$ for input size of $p$ and output size of $q$.
As the learning progresses, the noisy components are ignored in a state-conditional trend, giving the agent an implicitly driven exploration ability. 

Another important benefit of having NoisyNets is having access to a form of uncertainty estimation by using the predictive variance of a noisy layer as described in \cite{DBLP:journals/corr/abs-1812-02632}.
This is computed for action $a$ and state $s$ in the final layer that outputs $Q(s, a)$ values as follows:

\begin{equation}\label{noisy_pred_var}
\begin{aligned}
    Var[Q(s,a; \theta)] &= Var[w_a \phi(s)] + Var[b_a] \\
    & = \phi(s)^{\intercal} diag({\sigma_{w_a}}^2) \phi(s) + {\sigma_{b_a}}^2
\end{aligned}
\end{equation}

where $\phi(s)$ is the latent features generated from state $s$ to be fed this layer, $w_a$ and $b_a$ are the weight and bias terms, respectively.
The uncertainty in state $s$ is then obtained by taking the predictive variance of the action with largest Q-value as follows:

\begin{equation}\label{noisy_uncertainty}
    U(s) = Var[Q(s, \argmax{a}Q(s, a))]
\end{equation}

Even though our proposed method does not utilise uncertainty measurements, we employ NoisyNets in the student agent to make it possible to compare our approach with uncertainty-based ones as well.

\subsection{Random Network Distillation} \label{subsec:rnd} 

Random Network Distillation (RND) \cite{DBLP:journals/corr/abs-1810-12894} is proposed to address difficult exploration problems by providing the agent with an intrinsic state novelty bonus.
It uses two randomly initialised neural networks, \emph{target} and \emph{predictor}, denoted by $G$ and $\hat{G}$.
Over the course of learning, the \emph{predictor} network $\hat{G}$ is updated to minimise the mean squared error $\|\hat{G}(s) - G(s)\|^{2}$ between the embeddings of these two networks, for every state $s$ used to update the RL algorithm's model.
As the states are encountered in the learning, this error becomes smaller for such states as the predictor network becomes better at matching the target network's output.
Thus, this error term can be used as a measurement of novelty for any state.

\subsection{Action Advising} \label{subsec:action_advising}

Action advising \cite{DBLP:conf/atal/TorreyT13} is a form of inter-agent knowledge transfer method with a very simple mechanism. 
By using only a common understanding over a communication protocol and a set of actions, a teacher exchanges instructions in form of actions with a student agent in order to improve its learning process by biasing its interactions.
An important aspect of this framework is the limitation of the number in these interactions with a budget, considering the setbacks of real-world use cases.
For instance, the communication with the teacher may be costly, or its attention may be limited as in the case of a human teacher.
Furthermore, even with an available budget, the teacher can not be assumed to be available at all times.

While some of the approaches use simpler heuristics to distribute these advice, more advanced ones in deep RL employ techniques that involve estimating state novelty \cite{DBLP:conf/cig/IlhanGP19} or uncertainty \cite{DBLP:conf/aaai/SilvaHKT20a}\cite{DBLP:journals/corr/abs-1812-02632}  from the student agent's perspective.
Depending on the method, either of the peers can be initiating the knowledge transfer process. 
In this study, we follow the student-initiated approach and use the following baselines as well as state uncertainty and state novelty based ones:
\begin{itemize}
	\item \textbf{Early advising:} By intuition, the agents are thought to benefit more from teaching earlier in the learning process, which forms the main motivation of this heuristic \cite{DBLP:conf/atal/TorreyT13}.
	According to this, the peers request/receive advice constantly from the beginning until they run out of budget.
	\item \textbf{Random advising:} Despite the advantages of acquiring action advice early on, an agent may still discover some unfamiliar states later in the training as it learns.
	Therefore, it may also be helpful to distribute the available budget over a longer period of time.
	This heuristic provides an uninformed way of scattering these by making the peers determine to initiate action advising with some probability (usually $0.5$) at each timestep until the budget is consumed.
\end{itemize}

\section{The Game Environments} \label{sec:game_envs}

In this study, we employ two different game environments, namely, GridWorld and MinAtar, to represent different aspects of learning challenges.

\subsection{GridWorld} \label{subsec:gridworld}

\begin{figure}[!t]
\centering
\begin{subfigure}[b]{0.25\linewidth} 
    \centering
    \setlength{\fboxsep}{0pt}\fbox{\includegraphics[width=0.85\textwidth]{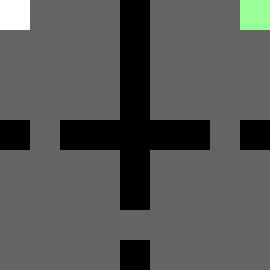}}
\end{subfigure}
\begin{subfigure}[b]{0.25\linewidth} 
    \centering
    \setlength{\fboxsep}{0pt}\fbox{\includegraphics[width=0.85\textwidth]{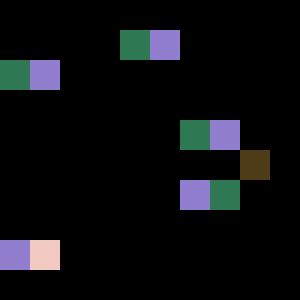}}
\end{subfigure}
\begin{subfigure}[b]{0.25\linewidth}
    \centering
    \setlength{\fboxsep}{0pt}\fbox{\includegraphics[width=0.85\textwidth]{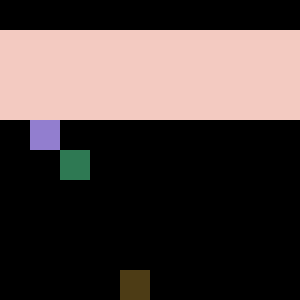}}
\end{subfigure}
\par\medskip
\begin{subfigure}[b]{0.25\linewidth}
    \centering
    \setlength{\fboxsep}{0pt}\fbox{\includegraphics[width=0.85\textwidth]{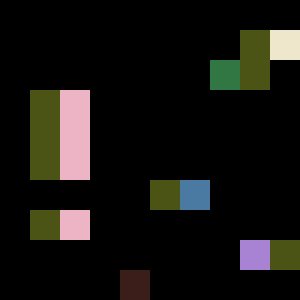}}
\end{subfigure}
\begin{subfigure}[b]{0.25\linewidth}
    \centering
    \setlength{\fboxsep}{0pt}\fbox{\includegraphics[width=0.85\textwidth]{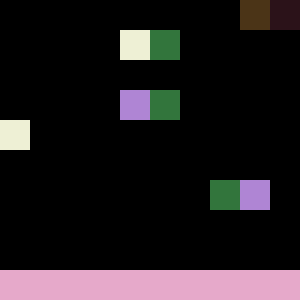}}
\end{subfigure}
\begin{subfigure}[b]{0.25\linewidth}
    \centering
    \setlength{\fboxsep}{0pt}\fbox{\includegraphics[width=0.85\textwidth]{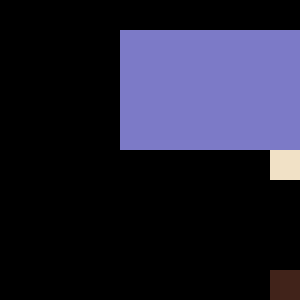}}
\end{subfigure}
\caption{Rendered observations of the initial state of GridWorld (top-left) and random states from MinAtar games Asterix (top-center), Breakout (top-right), Freeway (bottom-left), Seaquest (bottom-center) and Space Invaders (bottom-right).}
\label{fig_envs}
\end{figure}

GridWorld\footnote{\text{Codes are available at }\url{https://github.com/ercumentilhan/GridWorld}} is a grid structured environment with a size of $9 \times 9$, composed of ground (grey), pit (black) tiles, and goal (green) as visualised in Figure \ref{fig_envs} (left) where the agent is represented in white.
The environment is perceived by the agent as binary tensors with a size of $9\times9\times3$.
The agent starts in a fixed position in the top-left corner of the top-left room (as shown in a white cell in Figure \ref{fig_envs}, top-left), and must navigate to the goal tile in the top-right corner of the top-right room, with $4$ available actions in order to get a reward of $+1$.
Every time the agent attempts to take an action, it has a chance to execute another action uniformly at random instead, with a probability of $0.1$; however, if a random action happens to move the agent on a pit tile, it instead holds the agent in its previous position.
Reaching the goal, stepping on a pit tile or exceeding the maximum timesteps of $50$ terminates the episode with $0$ points of reward.
Considering the possibility of random movements and the tight timesteps limit, this game presents a significant exploration challenge despite its simple rules.

\subsection{MinAtar}

MinAtar \cite{DBLP:journals/corr/abs-1903-03176} is an environment composed of minimal versions of five popular Atari games\footnote{\text{Our version: }\url{https://github.com/ercumentilhan/MinAtar/tree/original}}.
While the core game mechanics are kept the same, the observation and the action spaces are reduced to cut down the representational complexity.
Every game has a common action space of $6$, and the observations are binary tensors with sizes of $10 \times 10 \times n$, where $n$ denotes the number of the object categories in the game.
Different from the default version, we set any game episode to have a maximum timesteps of $1000$.
Figure~\ref{fig_envs} shows screenshots taken from these games, and their brief descriptions are as follows:

\begin{itemize}
  \item \textbf{Asterix:} 
  The game has a constant stream of enemies and treasures that spawn and move in horizontal directions. 
  The agent can move in $4$ directions and must collect treasures to obtain $+1$ reward while avoiding enemies which kills the agent and terminates the game.
  
  \item \textbf{Breakout:}
  The agent controls a paddle at the bottom of the screen by moving it in horizontal directions.
  The objective is to keep hitting the ball to keep it within the screen to avoid losing the game.
  Hitting the bricks with the ball breaks them and yields $+1$ reward.
  New lines of bricks keep appearing after they are cleared up.
  
  \item \textbf{Freeway:}
  The agent's objective is to cross the way by avoiding the cars moving at different speeds in horizontal directions.
  Upon reaching the goal, $+1$ reward is awarded and the agent is sent back to the starting point, and the cars' directions and speeds are randomised.
  Being hit by a car also sends the agent back to the starting point.
  
  \item \textbf{Seaquest:}
  This is the game with the most complex rules among MinAtar games.
  The agent controls a submarine in a sea filled with enemy submarines, fish, and divers to be rescued by navigating around.
  Shooting the enemies yields $+1$ reward, as well as taking each diver to the surface.
  Moreover, the agent must keep an eye on the remaining oxygen level and have it replenished by going to the surface; which increases the difficulty each time it is done.
  If the agent goes up to the surface with no divers, is hit, or has no remaining oxygen, the game terminates.

  \item \textbf{Space Invaders:}
  The agent is in charge of controlling a space ship that can shoot bullets to the upcoming group of aliens from the top of the screen.
  Each shot down alien yields a reward of $+1$, and upon being cleared up, a new wave of aliens spawns with an increased movement speed.
  The aliens can also shoot back at the agent.
  If the agent is hit by a bullet or an alien, the game is terminated.

\end{itemize}

Directions of the moving sprites are also encoded in the observations by having a separate category for their trails to ensure full observability.
Asterix, Seaquest and Space Invaders also involve a periodical difficulty ramping that occurs at every $100^{th}$ timestep.

\section{Action Advising via Advice Novelty} \label{sec:proposed_method}

In our problem setting, we follow the standard RL framework and MDP formalisation as in Section~\ref{subsec:dqn}.
Our setup considers a situation where a student agent that employs an off-policy deep RL algorithm, e.g., DQN, with policy $\pi_{S}$ is learning to excel in a given task.
There is also a peer who is competent in this task to be treated as a teacher with a fixed policy $\pi_{T}$.
The student can access this peer and sample actions from $\pi_{T}$ for a limited number of times determined by the action advising budget $b$.
The objective of the agent is to utilise an action advising procedure to distribute the available advice requesting budget $b$ in the best possible way to maximise its learning performance.

The samples obtained in deep RL are not only used for discovering the environment but are also responsible for driving the learning of the agent's model. 
Additionally, every single step of environment interaction and model update influences the sample collection process right after. 
These make it very difficult for deep RL agents to predict what kind of long-term effects a transition, or a single piece of advice in this case, will have on their learning progress.
Therefore, student-initiated action advising techniques rely on simple heuristics and proxies of expected usefulness of samples in terms of learning contribution.
These, however, are prone to fail in several ways.

Early advising is a strong baseline due to the fact that the samples obtained early on have more influence on deep RL algorithms' learning progress.
However, it lacks the ability to distribute the available advising budget in more critical states.
This is likely to result in the budget being wasted, and make the agent miss important advice opportunities especially when the budget is small.
Additionally, having no stopping condition in making advice requests may cause the agent to get over-advised, which can deteriorate the learning performance as we show in Section~\ref{sec:results}.
Employing another common heuristic, uniformly random advising, can alleviate this drawback.
Nonetheless, not being able to follow the teacher's policy consistently causes this method to be unsuccessful in the tasks with sparse rewards that require deep exploration, as also shown in Section~\ref{sec:results}.

The more advanced techniques that rely on state importance surrogates such as novelty \cite{DBLP:conf/atal/SilvaGC17}\cite{DBLP:conf/cig/IlhanGP19}, model uncertainty \cite{DBLP:journals/corr/abs-1812-02632}\cite{DBLP:conf/aaai/SilvaHKT20a}, perform better in general.
Though, they still have drawbacks that can be problematic in some cases.

First of all, updates in these estimations are driven by the student's learning progression, regardless of having any teacher contribution in it.
If the teacher is present from the beginning, the student's RL model convergence can be assumed to be towards well-explored state-value targets with the support of teacher advice.
Otherwise, if the teacher remains absent for a significant amount of time as described in Section~\ref{sec:introduction}, there can be no guarantees for the student to explore the environment successfully.
As we show in Section~\ref{sec:results}, it is indeed very likely for the student to end up with a converged suboptimal policy even in a simple task with sparse rewards.
In this case, the student will ignore the teacher regardless of its potential knowledge to be leveraged.

Secondly, based on the student's deep RL model's properties, there may be some delays between the advice collection and its actual value to be taken effect in the model, e.g., off-policy updates with replay memory where the collected samples are held in a buffer and are employed periodically to update the model.
This lag prevents the student from being precise about what advice it actually needs.
Even though we do not demonstrate this behaviour in our experiments, we ensure there can be no model induced delays in estimations that affect our action advising approach.

Finally, these approaches require several restrictions on the student's RL algorithm.
On one hand, the state novelty-based approach needs to have access to the batches of samples the students use to update its learning model.
This prevents the student from remaining as a blackbox.
On the other hand, uncertainty-based methods require the student's model to be capable of providing a notion of uncertainty, which is possible only in a small subset of RL algorithms.

In order to address these shortcomings, we propose \emph{action advising via advice novelty}, a method that employs state novelty measurements to time advice requests.
In our technique, the student agent employs an RND module that consists of two randomly initialised neural networks $G$ and $\hat{G}$ with identical structures.
At each step with available advice requesting budget, the agent measures novelty $n_{s}$ of the state $s$ it encounters as the RND loss $n_{s} = \|\hat{G}(s) - G(s)\|^{2}$.
This value is then converted into a linearly decreasing advice requesting probability as $p_{s} = n_{s} / \eta $ (clipped to be in $[0, 1]$), where $\eta$ is a predefined threshold.
If the advice request takes place, then $\hat{G}$ is updated to minimise the loss term $\|\hat{G}(s) - G(s)\|^{2}$.
Thus, $n_{s}$ becomes smaller as the agent receives advice for $s$.
This can then be seen as a novelty metric for a piece of advice to be obtained in a particular state, considering the assumption of teacher policy $\pi_{T}$ being fixed, and ignoring the environment's stochasticity.
By performing updates only when advised, it is ensured that the student always attempts to learn from the teacher, no matter how far into convergence its task-level model is.
Furthermore, the RND updates occur right after the student is advised with only a single piece of observation rather than a batch of them.
This prevents the RND module to minimise its loss globally, and cause it to have relatively high novelty for the states it has not encountered in a while, giving these states a chance to be re-advised.
Finally, since RND employs neural networks with non-linear function approximation, our method can function in complex domains and is capable of generalising between unseen states.
A full description of our method can be seen in Algorithm \ref{alg:rm_regulation}.

\begin{algorithm}[tb]
\caption{Action Advising via Advice Novelty}
\label{alg:rm_regulation}
\begin{algorithmic}[1]
    \STATE {\bfseries Input:}     
    action advising budget $b$, agent policy $\pi_S$, teacher policy $\pi_{T}$, novelty threshold $\eta$
    \FOR{training steps $t \in \{1, 2, \ldots k\}$}
        \STATE get observation $s_t \sim Env$ if $Env$ is reset
        \STATE $a_t \leftarrow None$
        \IF{$b > 0$}
            \STATE $n_{s_t} \leftarrow \|\hat{G}(s_t) - G(s_t)\|^{2}$ \algorithmiccomment{compute novelty}
            \STATE $p_{s_t} \leftarrow n_{s_t} / \eta $ \algorithmiccomment{compute probability}  
            \STATE sample $p$ uniformly at random in $[0, 1]$
            \IF{$p_{s_t} > p$}
                \STATE $a_t \sim \pi_{T}$ \algorithmiccomment{request advice}
                \STATE Update $\hat{G}$ weights to minimise $\|\hat{G}(s) - G(s)\|^{2}$
                \STATE $b \leftarrow b - 1$
            \ENDIF
        \ENDIF 
        \IF{$a_t$ is $None$}
            \STATE $a_t \sim \pi_{S}$
        \ENDIF 
        \STATE Execute $a_t$ and obtain $r_t$, $s_{t+1} \sim Env$    
        \STATE Update the RL model, e.g., DQN.
        \STATE $s_t \leftarrow s_{t+1}$
    \ENDFOR
\end{algorithmic}
\end{algorithm}

\section{Experimental Setup} \label{sec:experimental_setup}

We are interested in investigating the shortcomings of the existing student-initiated action advising approaches, and evaluating how our approach compares with them in different scenarios.
For this purpose, we choose GridWorld and MinAtar games to conduct experiments in two stages involving different challenges. We compare the following modes of student agents with different action advising approaches:
\begin{itemize}
	\item \textbf{No Advising (None)}: The agent does not employ any form of action advising; it follows its own policy at all times.
	\item \textbf{Early Advising (EA)}: The agent follows the early advising heuristic to distribute its action advising budget by requesting advice until it runs out of its action advising budget.
	\item \textbf{Random Advising (RA)}: The agent follows the random advising heuristic and determines whether to request a piece of advice uniformly at random at each step until it runs out of its action advising budget.
	\item \textbf{Uncertainty-Based Advising (UA)}: Advice requests are made according to the student's RL model uncertainty, similarly to \cite{DBLP:journals/corr/abs-1812-02632, DBLP:conf/aaai/SilvaHKT20a}. 
	Specifically, at each step, the student's NoisyNets uncertainty is obtained for the current state, and then is divided by a threshold $\nu$ to determine advice requesting probability.
	\item \textbf{State Novelty-Based Advising (SNA)}: Advice requests are driven by state novelty measurements, similarly to \cite{DBLP:conf/cig/IlhanGP19}.
	For each state the student encounters, its novelty is measured by a separate RND module.
	Then, this value is divided by a predefined threshold $\rho$ to obtain advice requesting probability.
	The RND module is updated simultaneously with the student's RL model, by the same batches of samples.
	\item \textbf{Advice Novelty-Based Advising (ANA)}: The agent that follows our proposed approach.
\end{itemize}

\begin{table}[!t]
	\centering
	\caption{Hyperparameters of DQN (top section) and RND (bottom section).}
	\label{table:hyperparameters}
	\begin{tabular}{l|c|c}  	
	     & \multicolumn{2}{c}{Value}  \\
		Hyperparameter name & GridWorld & MinAtar \\
  	    \cmidrule(r){1-3}
  	    Replay memory size to start learning & $1000$ & $10000$ \\
		Replay memory capacity & $10000$ & $100000$ \\
		Target network update period & $250$ & $1000$ \\
		Minibatch size & $32$ & $32$ \\
		Learning rate & $0.0001$ & $0.0001$ \\
		Train period & $2$ & $2$ \\
		Huber loss $\delta$ & $1$ & $1$ \\
		\cmidrule(r){1-3}
		Learning rate & $0.0001$ & $0.0001$ \\
		Normalisation steps & $1$k & $5$k \\	
	\end{tabular}
\end{table}

The student agent's RL algorithm is set to be DQN, including the prominent extensions of double Q-learning, dueling networks, and NoisyNets exploration strategy.
The neural network structure is comprised of a single convolutional layer consisting $16$ $3 \times 3$ filters with a stride of $1$, followed by a fully connected noisy layer with $128$ hidden units.
The RND module employs a similar neural network structure with a single convolutional layer consisting $16$ $3 \times 3$ filters with a stride of $1$, followed by a regular fully connected layer with $128$ hidden units and $6$ output units (set arbitrarily).
These networks do not share any weights or sample batches used during the training.
The hyperparameters of DQN and RND are tuned for each environment prior to the experiments (can be seen in Table \ref{table:hyperparameters}) and the discount factor is set as $0.99$ in all experiments.

As described in the problem setup, there needs to be a teacher for the student agent to be able to get advice from.
In GridWorld, we set our teacher policy as following the shortest path from the current position to the goal tile.
In MinAtar, we generated competent teachers by training separate DQN agents for each of the games for $3$M steps, who achieve final evaluation scores (as defined later on) of $29.28$, $81.15$, $5.77$, $146.64$, $146.06$ in Asterix, Breakout, Freeway, Seaquest, Space Invaders, respectively.
The teachers are made to employ $\epsilon$-greedy exploration instead of NoisyNets, to have them as dissimilar as possible from the student in order to eliminate any possible advantages that may arise in the knowledge exchange process due to them being identical.

Every student variant (e.g., None, EA, etc.) is trained for a fixed number of steps which we define as a learning session.
During a learning session, the agent is evaluated periodically in a separate sequence of episodes with having any form of exploration (e.g., the noise perturbations of NoisyNets) and teaching disabled.
The scores obtained in these episodes are averaged to determine the evaluation score for this evaluation step.
These scores reflect the agent's actual expertise in the corresponding step in the learning session.
In GridWorld, since the actual cumulative rewards at the end of an episode is either $0$ or $1$, a more informative evaluation score is defined to be in $[0, 1]$ by taking the number of remaining timesteps and distance to the goal tile into calculations.
In MinAtar, the original game scores in the framework are used as evaluation scores.
In addition to the evaluation scores, we also plot the number of advice taken in every $100$ steps as well as cumulatively, to observe the trends in budget spending.

In the first stage of experiments, we use GridWorld as a simple and interpretable task to highlight the aforementioned drawbacks, as well as performing a preliminary benchmark on the methods to validate their suitability for tasks with more complex mechanics.
The learning session lengths are set as $100$k steps, and evaluations are performed at every $100^{th}$ step in a sequence of $5$ episodes.
We set two different scenarios, namely, Scenario I and Scenario II, with small and large budget options of $5$k and $50$k each, resulting in $4$ cases in total.
In Scenario I, the teacher is present from the beginning of learning sessions, which is the common experimental setting used in the previous action advising studies.
In Scenario II, the teacher joins the loop at the $25$k$^{th}$ step.
By having such a scenario, we test the ability of the student in dealing with the belated teacher.
UA, SNA, ANA hyperparameters $\nu$, $\rho$, $\eta$ are determined empirically in the $5$k budget setting, to be $0.001$, $0.0001$, $0.001$ and are kept the same for all $4$ cases.

In the second stage, we evaluate the approaches in a set of tasks with more complex dynamics presented via $5$ different games in MinAtar environment. 
Learning sessions are set to have a length of $1.5$M steps and evaluations are performed at every $1000^{th}$ step in a sequence of $5$ episodes.
Since the Scenario II experiments in GridWorld are sufficient to demonstrate the weakness regarding the extensive unavailability of the teacher, experiments in MinAtar are only conducted for Scenario I to evaluate the general performance of the methods, again with two different budget options of $50$k and $250$k.
UA, SNA, ANA hyperparameters $\nu$, $\rho$, $\eta$ are determined empirically on game by game basis for $50$k budget, to be $0.0001$, $0.0025$, $0.001$ for Asterix; $0.001$, $0.025$, $0.025$ for Breakout; $0.0001$, $0.1$, $0.05$ for Freeway; $0.001$, $0.05$, $0.05$ for Seaquest; $0.0025$, $0.01$, $0.0025$ for Space Invaders, respectively.

Experiment results are aggregated over $9$ different seeds for GridWorld, and over $8$ different seeds for MinAtar games.
Training and evaluation episode sequences (random game events) are fixed via random seeds and kept the same across different experiment seeds.
Therefore, these experiment seeds only affect the agents' internal computations.
Finally, we perform RND observation normalisation as in \cite{DBLP:journals/corr/abs-1810-12894} by using the mean and standard deviation calculated over the first $1000$ and $5000$ observations, in GridWorld and MinAtar, respectively\footnote{\text{Codes for our experiments can be found at }\url{https://github.com/ercumentilhan/advice-novelty}}.

\section{Results and Discussion} \label{sec:results}

\begin{figure*}[!t]
\centering
\includegraphics[width=1.0\textwidth]{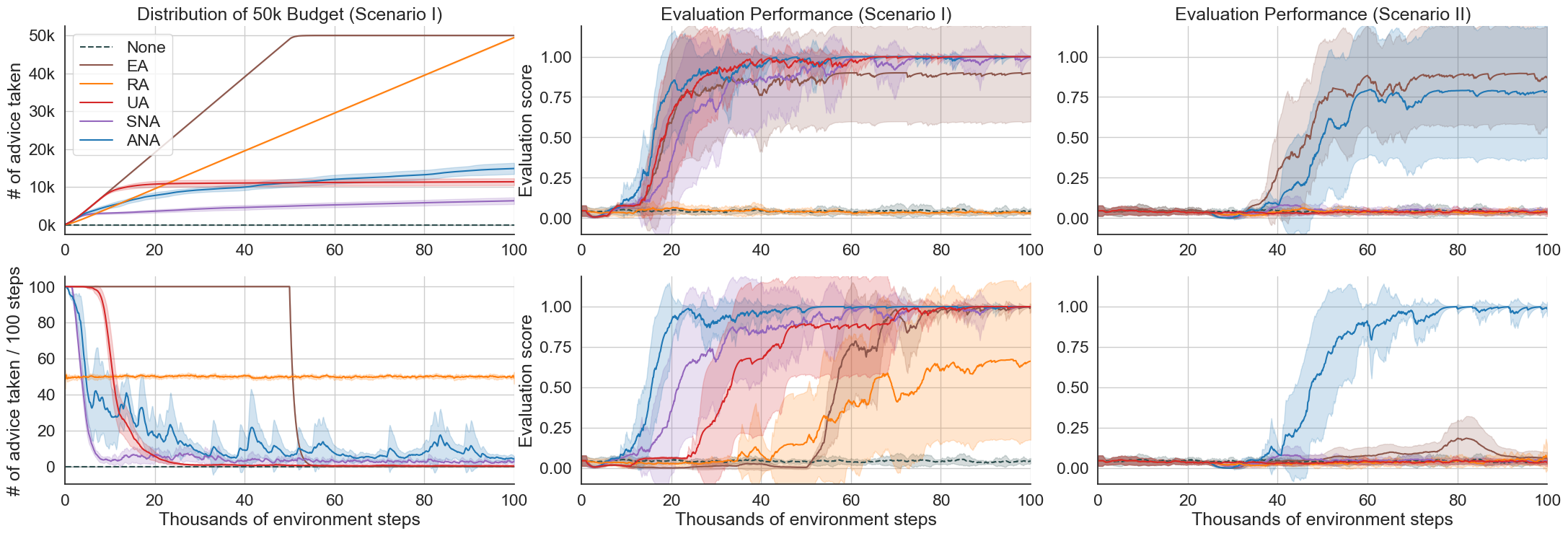}
\caption{
Number of advice taken cumulatively and in every $100$ steps period in Scenario I with $50$k budget (leftmost column), evaluation scores in Scenario I (middle column) and Scenario II (rightmost column) with $5$k (top row) and $50$k (bottom row) budgets, obtained in GridWorld game with no action advising (None) and action advising methods EA, RA, UA, SNA, ANA.
}
\label{fig:gridworld_results}
\end{figure*}

\begin{figure*}[!t]
\centering
\includegraphics[width=1.0\textwidth]{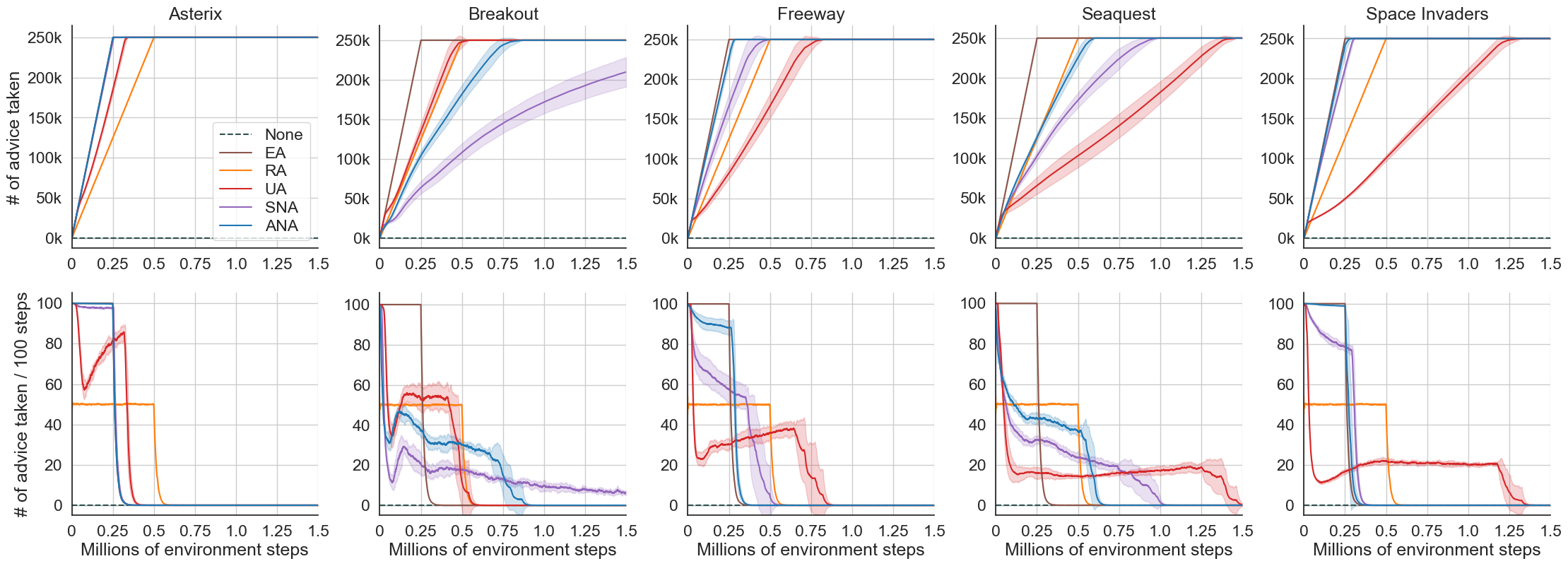}
\caption{
Number of advice taken cumulatively and in every $100$ steps period in Scenario I with $250$k budget, obtained in five MinAtar games with no action advising (None) and action advising methods EA, RA, UA, SNA, ANA.
}
\label{fig:minatar_budget_results}
\end{figure*}

\begin{figure*}[!t]
\centering
\includegraphics[width=1.0\textwidth]{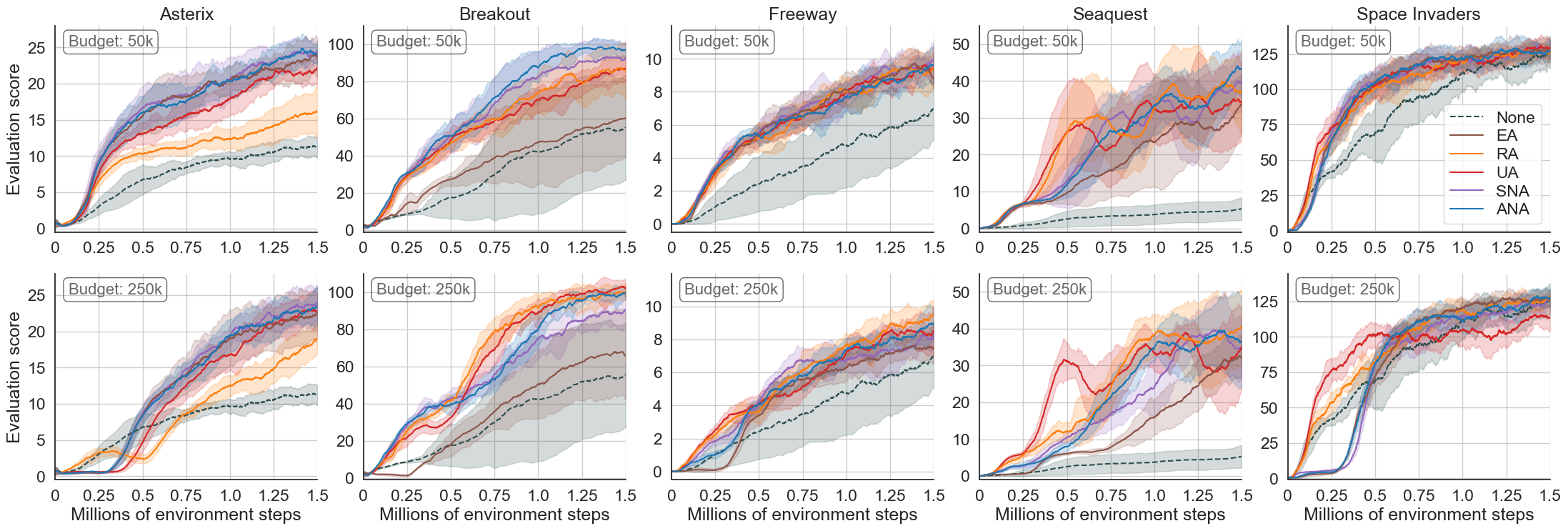}
\caption{
Evaluation scores of Scenario I with $50$k (top row) and $250$k (bottom row) budgets, obtained in five MinAtar games with no action advising (None) and action advising methods EA, RA, UA, SNA, ANA.
}
\label{fig:minatar_score_results}
\end{figure*}

\begin{table*}[t]
	\centering
	\caption{
	Area under the curve (AUC) and the mean of last $50$ values (Final Score) of the evaluation score plots of None, EA, RA, UA, SNA, ANA agent modes obtained in five MinAtar games averaged over $5$ runs.
	The numbers denoted by $\pm$ represent standard deviation. 
	The percentage values in parentheses indicate the relative difference to the values obtained by EA.
	\emph{Overall} section in bottom presents these percentages averaged over these $5$ games.
	$(+)$ and $(-)$ on the left-hand side of the results indicate whether ANA's results are significantly better or worse than them, respectively (according to Welch's $t$-test with $p < 0.05$).
	The best results in their own brackets are denoted in bold.
	}
	\label{table:minatar_results}
	\begin{tabular}{P{0.8cm}p{0.4cm}P{0.11cm}p{3cm}p{0.11cm}p{3.2cm}p{0.11cm}p{3cm}p{0.11cm}p{3.2cm}}  
	\toprule
	& & \multicolumn{4}{c}{Final Score} & \multicolumn{4}{c}{AUC ($\times 10^3$)} \\
	\cmidrule(r){3-6}
	\cmidrule(r){7-10}
	Game & Mode & \multicolumn{2}{c}{$50$k} & \multicolumn{2}{c}{$250$k} & \multicolumn{2}{c}{$50$k} & \multicolumn{2}{c}{$250$k} \\	
	\midrule
	
	\multirow{6}{*}{Asterix} &
	None & 	
	$(+)$ & $11.29 \pm 1.3$ $(-52.4\%)$ & 
	$(+)$ & $11.29 \pm 1.3$ $(-49.0\%)$ &
	$(+)$ & $11.04 \pm 0.7$ $(-54.4\%)$ &
	$(+)$ & $11.04 \pm 0.7$ $(-40.2\%)$ 
	\\
   	& EA &
   	& $23.74 \pm 2.3$ & 
   	& $22.16 \pm 3.1$ &
   	& $24.20 \pm 2.4$ & 
   	& $18.45 \pm 2.8$ 
    \\
	& RA &   	
	$(+)$ & $15.98 \pm 2.8$ $(-32.7\%)$ & 
	$(+)$ & $11.21 \pm 1.5$ $(-46.8\%)$ &
	 $(+)$ & $15.75 \pm 1.4$ $(-34.9\%)$ &
	$(+)$ & $11.01 \pm 0.9$ $(-39.1\%)$ 
    \\
	& UA &
	 $(+)$ & $21.74 \pm 1.6$ $(-8.4\%)$ & 
	$(+)$ & $11.21 \pm 1.5$ $(-46.8\%)$ &
	$(+)$ & $21.39 \pm 1.5$ $(-11.6\%)$ &
	$(+)$ & $11.01 \pm 0.9$ $(-39.1\%)$ 
    \\
	& SNA &
	& \bm{$24.26 \pm 1.5$} \bm{$(+2.2\%)$} & 
	 & \bm{$23.69 \pm 2.3$} \bm{$(+6.9\%)$} &
	& \bm{$24.97 \pm 2.7$} \bm{$(+3.2\%)$} &
	& \bm{$19.19 \pm 1.4$} \bm{$(+4.0\%)$}
    \\
	& ANA & 
	& \bm{$24.26 \pm 1.2$} \bm{$(+2.2\%)$} & 
	& $23.18 \pm 2.1$ $(+4.6\%)$ &
	& $24.38 \pm 1.8$ $(+0.7\%)$ &
	& $18.74 \pm 1.6$ $(+1.6\%)$
    \\
    
    \cmidrule(r){1-10}

	\multirow{6}{*}{Breakout} &
	None & 
	$(+)$ & $54.12 \pm 26.4$ $(-9.5\%)$ & 
	$(+)$ & $54.12 \pm 26.4$ $(-19.4\%)$ &
	$(+)$ & $44.47 \pm 25.7$ $(-15.4\%)$ &
	$(+)$ & $44.47 \pm 25.7$ $(-12.8\%)$
	\\
   	& EA &
	$(+)$ & $59.81 \pm 20.2$ & 
	$(+)$ & $67.17 \pm 21.7$  &
	$(+)$ & $52.53 \pm 16.7$  &
	$(+)$ & $51.00 \pm 13.9$ 
    \\
	& RA &
	& $86.78 \pm 12.8$ $(+45.1\%)$ & 
	& $99.81 \pm 3.7$ $(+48.6\%)$ &
	$(+)$ & $83.77 \pm 8.3$ $(+59.5\%)$ &
	$(-)$ & \bm{$96.02 \pm 3.9$} \bm{$(+88.3\%)$}
    \\
	& UA &
	$(+)$ & $86.68 \pm 6.3$ $(+44.9\%)$ & 
	& \bm{$102.42 \pm 3.7$} \bm{$(+52.5\%)$} &
	$(+)$ & $81.88 \pm 3.8$ $(+55.9\%)$ &
	$(-)$ & $92.21 \pm 3.0$ $(+80.8\%)$ 
    \\
	& SNA &
	& $92.12 \pm 8.1$ $(+54.0\%)$ & 
	$(+)$ & $89.04 \pm 8.1$ $(+32.6\%)$ &
    & $91.86 \pm 10.7$ $(+74.9\%)$	 &
	& $82.20 \pm 6.2$ $(+61.2\%)$
    \\
	& ANA &
	& \bm{$97.18 \pm 3.6$} \bm{$(+62.5\%)$} & 
	& $98.62 \pm 4.0$ $(+46.8\%)$ &
    & \bm{$97.12 \pm 6.3$} \bm{$(+84.9\%)$} &
	& $87.65 \pm 3.7$ $(+71.9\%)$
    \\ 
    \cmidrule(r){1-10} 
	\multirow{6}{*}{Freeway} &
	None &
	$(+)$ & $6.71 \pm 1.8$ $(-26.7\%)$ & 
	$(+)$ & $6.71 \pm 1.8$ $(-10.5\%)$ &
	$(+)$ & $5.18 \pm 2.3$ $(-43.7\%)$ &
	$(+)$ & $5.18 \pm 2.3$ $(-21.5\%)$ 
	\\
   	& EA & 
	 & $9.15 \pm 0.5$ & 
	$(+)$ & $7.49 \pm 0.7$&
	& $9.21 \pm 0.7$ &
	$(+)$ & $6.61 \pm 0.5$
    \\
	& RA &
	& $9.22 \pm 1.2$ $(+0.8\%)$ & 
	& \bm{$9.31 \pm 0.8$} \bm{$(+24.2\%)$} &
	& $9.07 \pm 0.5$ $(-1.6\%)$ &
	$(-)$ & \bm{$8.29 \pm 0.6$} \bm{$(+25.5\%)$} 
    \\
	& UA &
	& $9.45 \pm 0.7$ $(+3.3\%)$ & 
    & $8.32 \pm 1.2$ $(+11.1\%)$ &
	& $9.10 \pm 0.3$ $(-1.3\%)$ &
	& $7.84 \pm 0.5$ $(+18.7\%)$
    \\
	& SNA &
	& \bm{$9.76 \pm 0.8$} \bm{$(+6.6\%)$} & 
	& $8.09 \pm 0.6$ $(+8.0\%)$ &
	& \bm{$9.45 \pm 0.6$} \bm{$(+2.5\%)$} &
	& $7.70 \pm 0.7$ $(+16.6\%)$
    \\
	& ANA & 
	& $9.52 \pm 0.9$ $(+4.1\%)$ & 
	 & $8.86 \pm 1.0$ $(+18.3\%)$ &
	& $9.04 \pm 0.5$ $(-1.9\%)$ &
	& $7.62 \pm 0.4$ $(+15.3\%)$
    \\
    \cmidrule(r){1-10}
	\multirow{6}{*}{Seaquest} &
	None &
	$(+)$ & $5.13 \pm 2.7$ $(-84.1\%)$ & 
	$(+)$ & $5.13 \pm 2.7$ $(-84.0\%)$ &
	$(+)$ & $4.34 \pm 2.7$ $(-82.5\%)$ &
	$(+)$ & $4.34 \pm 2.7$ $(-75.9\%)$  
	\\
   	& EA &
	$(+)$ & $32.25 \pm 8.0$ & 
	& $32.07 \pm 5.0$&
	$(+)$ & $24.83 \pm 6.9$&
	$(+)$ & $18.04 \pm 3.0$
    \\
	& RA & 
	& $37.26 \pm 8.8$ $(+15.5\%)$ & 
	& \bm{$39.68 \pm 9.4$} \bm{$(+23.7\%)$} &
	& \bm{$36.73 \pm 8.0$} \bm{$(+48.0\%)$} &
	& $34.38 \pm 4.3$ $(+90.5\%)$ 
    \\
	& UA &
	& $34.68 \pm 10.9$ $(+7.5\%)$ & 
	& $33.31 \pm 11.6$ $(+3.9\%)$ &
	& $35.05 \pm 5.6$ $(+41.2\%)$ &
	$(-)$ & \bm{$36.21 \pm 2.0$} \bm{$(+100.7\%)$} 
    \\
	& SNA & 
	& $39.26 \pm 7.1$ $(+21.7\%)$ & 
	& $34.74 \pm 7.6$ $(+8.4\%)$ &
	& $32.78 \pm 3.5$ $(+32.0\%)$ &
	& $28.00 \pm 2.6$ $(+55.2\%)$ 
    \\
	& ANA &
	& \bm{$43.23 \pm 6.6$} \bm{$(+34.0\%)$} & 
	& $36.95 \pm 11.7$ $(+15.2\%)$ &
	& $32.95 \pm 4.7$ $(+32.7\%)$ &
	& $30.50 \pm 2.4$ $(+69.0\%)$ 
    \\
    \cmidrule(r){1-10}
	\multirow{6}{*}{\shortstack{Space\\Invaders}} &
	None &
	& $124.51 \pm 9.2$ $(-2.3\%)$ & 
	& $124.51 \pm 9.2$ $(-2.0\%)$ &
	$(+)$ & $125.12 \pm 13.1$ $(-15.8\%)$ &
	& $125.12 \pm 13.1$ $(+0.8\%)$ 
	\\
   	& EA &
	& $127.48 \pm 8.3$ & 
	& $127.05 \pm 5.5$ &
	& $148.59 \pm 7.0$ &
	& $124.13 \pm 3.9$
    \\
	& RA &
	& $126.66 \pm 4.4$ $(-0.6\%)$ & 
	& $126.46 \pm 3.7$ $(-0.5\%)$ &
	& $146.55 \pm 6.7$ $(-1.4\%)$ &
	$(-)$ & \bm{$137.80 \pm 4.2$} \bm{$(+11.0\%)$}
    \\
	& UA &
	& \bm{$129.19 \pm 6.4$} \bm{$(+1.3\%)$} & 
	$(+)$ & $114.36 \pm 8.1$ $(-10.0\%)$ &
    & \bm{$149.92 \pm 7.0$} \bm{$(+0.9\%)$} &
	$(-)$ & $135.29 \pm 6.0$ $(+9.0\%)$
    \\
	& SNA & 
	& $127.70 \pm 3.4$ $(+0.2\%)$ & 
	& $124.58 \pm 4.7$ $(-1.9\%)$ &
	& $148.85 \pm 5.7$ $(+0.2\%)$ &
	& $118.43 \pm 7.1$ $(-4.6\%)$  
    \\
	& ANA &
	& $126.05 \pm 5.1$ $(-1.1\%)$ & 
	& \bm{$127.77 \pm 6.0$} \bm{$(+0.6\%)$} &
	& $148.41 \pm 8.2$ $(-0.1\%)$ &
	& $123.16 \pm 3.4$ $(-0.8\%)$ 
    \\
    \cmidrule(r){1-10}
	\multirow{5}{*}{\emph{Overall}} &
	None &  
	& \multicolumn{1}{c}{$-35.0\%$} & 
	& \multicolumn{1}{c}{$-33.0\%$} &
	& \multicolumn{1}{c}{$-42.4\%$} & 
	& \multicolumn{1}{c}{$-29.9\%$} 
	\\
	& RA & 
	& \multicolumn{1}{c}{$+5.6\%$} & 
	& \multicolumn{1}{c}{$+15.9\%$} &
	& \multicolumn{1}{c}{$+13.9\%$} & 
	& \multicolumn{1}{c}{$+37.0\%$} 
    \\
	& UA &
	& \multicolumn{1}{c}{$+9.7\%$} & 
	& \multicolumn{1}{c}{$+12.1\%$} &
	& \multicolumn{1}{c}{$+17.0\%$} & 
	& \multicolumn{1}{c}{$\bm{+40.1\%}$} 
    \\
	& SNA &
	& \multicolumn{1}{c}{$+17.0\%$} & 
	& \multicolumn{1}{c}{$+10.8\%$} &
	& \multicolumn{1}{c}{$+22.5\%$} & 
	& \multicolumn{1}{c}{$+26.5\%$} 
    \\
	& ANA &
	& \multicolumn{1}{c}{$\bm{+20.3\%}$} & 
	& \multicolumn{1}{c}{$\bm{+17.1\%}$} &
	& \multicolumn{1}{c}{$\bm{+23.3\%}$} & 
	& \multicolumn{1}{c}{$+31.4\%$} 
    \\

	\bottomrule
	\end{tabular}
\end{table*}

The results of our experiments are presented in Figure~\ref{fig:gridworld_results} (GridWorld); and in Figures~\ref{fig:minatar_budget_results} and \ref{fig:minatar_score_results}, and Table~\ref{table:minatar_results} (MinAtar). 
In Figure~\ref{fig:gridworld_results}, the leftmost column contains two plots for the number of advice taken in total and in every $100$ steps, in Scenario I with the budget amount of $50$k; the middle column displays the evaluation scores in Scenario I, and the rightmost column displays the evaluation scores in Scenario II (with the budgets of $5$k on top and $50$k on bottom).
Figure~\ref{fig:minatar_budget_results} includes the plots of the number of advice taken in total and in every $100$ steps, in Scenario I with the budget amount of $250$k for all five MinAtar games.
In Figure~\ref{fig:minatar_budget_results}, plots of evaluation scores obtained in MinAtar games with two different budget settings of $50$k (top row) and $250$k (bottom row) are displayed.
These results are presented numerically in Table~\ref{table:minatar_results} with the area under the curve and final values. 
The table also shows whether the results are significantly different than ANA's results according to Welch's $t$-test with $p < 0.05$.
$(+)$ sign on the left-hand side of a result indicates that ANA is significantly better than the corresponding method.
Similarly, $(-)$ indicates that ANA is significantly worse than it.
We also denoted the best results in their own brackets in bold.
The plots in Figure~\ref{fig:gridworld_results} and Figure~\ref{fig:minatar_budget_results} that display the number of advice taken are only generated for Scenario I with the maximum budget settings since the budget distribution is identical in the cases with smaller budgets with only the difference of being cut off early.
The curves are plotted with appropriate moving average smoothing for the sake of comprehensibility, and the standard deviation across the runs are shown with the shaded areas.

\subsection{GridWorld} \label{sec:results_gridworld}

In Scenario I with a small budget of $5$k, all of the action advising methods performed reasonably well, with the exception of RA which fails to be any better than None.
The poor performance of None indicates how challenging it can be to conduct exploration successfully even with an advanced method like NoisyNets when the time constraints are tight as in this GridWorld game. 
Despite taking plenty of expert advice, RA also fails due to its inability to consistently follow the teacher, which is especially essential in the tasks requiring deep exploration like this one.
This makes RA an unreliable action advising heuristic.

When the budget is increased to $50$k, we see how the performances collapse due to taking too much advice hence not executing their own policies adequately to collect integral samples.
This is most obvious in EA since it employs the most greedy way of spending the budget amongst all methods, which makes it dangerously susceptible to such high budget settings.
With the addition of more budget, RA finally manages to benefit from expert advice, however still very inadequately.
The advanced methods of UA, SNA, and our ANA do a good job on not overusing all the budget given to them even though they are not tuned to handle $50$K.
While UA performs slightly worse, ANA does better with the addition of an extra budget.
As it can be seen, ANA does this while using more budget in the end than UA; this shows that it is not just about cutting off the advice requests but is also about distributing them in the appropriate states and across the learning session.
In terms of budget efficiency, SNA seems to be doing the best in this case; however, also has worse task performance.
Finally, differently than UA and SNA, ANA is observed to follow a trend with occasional peaks in the number of advice taken per $100$ steps.
This is very likely to be caused by its RND update rule that uses single samples rather than batches that are also non-i.i.d. when the advice requests are made consecutively. 
As a result, RND model does not achieve global optimum and remain to yield significantly higher loss for the sample(s) that are not encountered recently.
This is a unique characteristic of ANA which can be advantageous as it makes the teacher advice to be re-acquired occasionally.

In Scenario II, where the teacher joins the learning session at $25$k$^{th}$ step, both UA and SNA fail to learn from the teacher as expected, due to their teacher independent convergence in estimations of uncertainty and novelty, respectively.
Our method ANA, however, manages to leverage the teacher's knowledge in both budget options, despite of the student being converged to suboptimal Q-value targets due to under-exploration.
Clearly, if there is such a possibility as depicted in this scenario, methods like UA and SNA are not going to be suitable action advising methods to be employed.

\subsection{MinAtar} \label{sec:results_minatar}

Results in MinAtar games represent a more general performance evaluation of action advising techniques in a variety of tasks with commonly experimented settings.
Even though the final performance is our main concern, we also present AUC values to give an idea regarding the methods' learning accelerations.

We first discuss the results obtained with a budget of $50$k as the main comparison case since it is the setting the tested methods are tuned for.
In Asterix, ANA and SNA are the best performing methods with EA being very close to them, and it is followed by UA and RA.
Considering how RA fails and the successful budget spending patterns of EA, SNA and ANA; it can be speculated that is critical in this particular game to follow the teacher over many consecutive steps, similarly to GridWorld.
In Breakout, ANA outperforms all other methods which already do well compared to None and EA.
This in comparison to the case in Asterix shows how EA and RA can be unreliable choices as a trade-off for being very simple heuristics.
Unlike GridWorld and Asterix, the successful methods in Breakout are the ones that ask for advice less frequently; this may be due to Breakout not requiring expert advice over many steps since the game events unfold on their own once the ball is hit with the paddle, and the different random moves taken in the meantime may be much more valuable sources to reduce RL model error, rather than taking the same expert advised actions such as just waiting stationarily until the ball traverses back down.
Seaquest has very interesting results where every method achieves different standings in different stages of the learning session.
In terms of the final performance, however, ANA manages to come on top again with SNA and RA following it after and UA performing rather poorly despite its rapid progression earlier in learning.
As a result of Seaquest mechanics being the most complicated amongst all, it is not very clear what causes the learning fluctuations in these plots.
In Freeway and Space Invaders, despite all the different advice requesting patterns followed by different methods, they are very similar when it comes to the learning performance.
Even though it may be surprising at first considering that Freeway is a game with sparse rewards, its action space and the possible positions the agent can traverse in the game grid spatially are rather small.
Therefore, there is almost no need for the expert advice to be distributed strategically across the game episode.

When the budget is increased to $250$k, we observe a fair amount of performance deterioration, especially in EA, SNA and ANA, which is especially visible in the learning progress.
In Asterix, Freeway, and Space Invaders, even though the final performance is not affected greatly, there is a sharp drop in learning progression caused by the over-advising induced delay in the collection of useful samples.
This is closely linked to them behaving more similarly to EA in these cases as it can be seen in the budget plots.
These results emphasise the importance of handling redundant advice budgets.
Currently, none of these action advising methods are aware of how many times they will get to ask for advice.
Instead, they are designed to make the most out of some supposedly small budget they are given, without any notion of long-term planning of its utilisation.

Overall, as the performance superiority of all the methods over None suggests, action advising can provide substantial advantages to accelerate learning.
Our ANA is the winner in terms of the final performance both in $50$k and $250$k budget options.
The significance analyses also show ANA provides a clear advantage over its alternatives.
Depending on the budget, ANA's performance is followed by methods in a different order.
For instance, while SNA is far ahead of others and is slightly behind ANA in $50$k budget, RA takes its place in the $250$k budget scenario.
The decline in performance with higher budgets and RA's robustness to this by spacing out the advice requests to allow the student to execute its self policy more often points out the importance of collecting on-policy samples adequately even when the agent itself employs an off-policy RL algorithm, as highlighted in \cite{DBLP:conf/icml/FujimotoMP19}\cite{DBLP:conf/nips/KumarFSTL19} as well.
Clearly, different games require different strategies to distribute the action advising budget, and there is no straightforward way to determine a way that applies to all cases successfully.
Furthermore, the ways these methods behave are also dependent on the underlying task, since the uncertainty and novelty estimations may be affected differently even with the identical streams of observations.
For instance, while UA tends to spend its budget more slowly than others in most cases, in Breakout it behaves differently to output its best possible performance.
Finally, it is worth it to mention that ANA is also observed to stand out in terms of computational efficiency compared to the runner-up SNA; while ANA updates its RND model with a single sample only when it successfully receives advice, SNA updates it for a batch of samples every time it performs learning until its budget reaches zero.

\section{Conclusions} \label{sec:conclusions}

In this work, we evaluated the prominent student-initiated action advising methods that are compatible with off-policy deep RL agents; and highlighted their shortcomings such as not being able to handle the belated availability of the teacher, or requiring very specific underlying algorithms to work.
To address these, we proposed an alternative student-initiated action advising algorithm that utilises state novelty computed via Random Network Distillation (RND) to determine when to request a piece of advice.
Differently from the previous works, RND is updated only with the states that are involved in the advice exchanges.
Thus, it is ensured that the student will take advantage of the teacher as soon as it becomes available regardless of its own RL model's convergence.

Empirical results in GridWorld and MinAtar games validate our speculations of the aforementioned drawbacks and show that the state-of-the-art methods that utilise state novelty or uncertainty can be ineffective if the teacher is not present from the beginning.
Furthermore, our advice novelty approach manages to be a very reliable choice by yielding the best final performance in the majority of these experiments, as well as being able to handle belated teachers, not requiring particular models to access uncertainty estimations, and also not interfering with the student's RL algorithm.
It is also seen that there is no trivial way to define a general action advising strategy to distribute the budget efficiently across many different cases.
Finally, it is found to be challenging for even the most complicated methods to handle excessive budgets without encountering a significant performance deterioration.
Accordingly, the hyperparameters that are responsible to manage budget distribution require careful tuning considering both the task characteristics and the total available budget.

An interesting extension of this study would be further investigating the components of our approach to precisely determine how they behave and how their variants affect the performance.
For instance, training RND incrementally with batches drawn from the complete set of collected advice instead of only the latest samples can provide valuable insights for comparison.
Another direction for future work could involve devising a form of threshold adaptation to make the action advising techniques more robust against the changes in task and budget specifications.
Additionally, further analyses in the dynamics of different RL algorithms operating with action advising would be imperative to invent more general action advising methods.
Our study employs a student agent with DQN and NoisyNets exploration; it will be worthwhile to investigate the performance of the action advising algorithms with different RL algorithms and exploration methods to see how their standings vary.
Finally, there is still a significant research gap in action advising with multiple teachers which requires further attention, and we believe that our approach is likely to be useful in such a problem setting.

\section*{Acknowledgment}
This research utilised Queen Mary's Apocrita HPC facility, supported by QMUL Research-IT. http://doi.org/10.5281/zenodo.438045

\printbibliography

@article{DBLP:journals/corr/abs-1708-05866,
  author    = {Kai Arulkumaran and
               Marc Peter Deisenroth and
               Miles Brundage and
               Anil Anthony Bharath},
  title     = {A Brief Survey of Deep Reinforcement Learning},
  journal   = {CoRR},
  volume    = {abs/1708.05866},
  year      = {2017}
}

@article{DBLP:journals/corr/abs-1812-02632,
  author    = {Si{-}An Chen and
               Voot Tangkaratt and
               Hsuan{-}Tien Lin and
               Masashi Sugiyama},
  title     = {Active Deep Q-learning with Demonstration},
  journal   = {CoRR},
  volume    = {abs/1812.02632},
  year      = {2018}
}

@techreport{settles2009active,
  title={Active learning literature survey},
  author={Settles, Burr},
  year={2009},
  institution={University of Wisconsin-Madison Department of Computer Sciences}
}

@article{DBLP:journals/aamas/SilvaWCS20,
  author    = {Felipe Leno da Silva and
               Garrett Warnell and
               Anna Helena Reali Costa and
               Peter Stone},
  title     = {Agents teaching agents: a survey on inter-agent transfer learning},
  journal   = {Auton. Agents Multi Agent Syst.},
  volume    = {34},
  number    = {1},
  pages     = {9},
  year      = {2020}
}

@misc{alphastarblog,
  title="{AlphaStar: Mastering the Real-Time Strategy Game StarCraft II}",
  author={Vinyals, Oriol and Babuschkin, Igor and Chung, Junyoung and others},
  howpublished={\url{https://deepmind.com/blog/alphastar-mastering-real-time-strategy-game-starcraft-ii/}},
  year={2019}
}

@article{DBLP:journals/corr/abs-1908-02388,
  author    = {Adrien Ali Ta{\"{\i}}ga and
               William Fedus and
               Marlos C. Machado and others},
  title     = {Benchmarking Bonus-Based Exploration Methods on the Arcade Learning
               Environment},
  journal   = {CoRR},
  volume    = {abs/1908.02388},
  year      = {2019}
}

@inproceedings{DBLP:conf/aaai/HesterVPLSPHQSO18,
  author    = {Todd Hester and
               Matej Vecer{\'{\i}}k and
               Olivier Pietquin and others},
  title     = {Deep Q-learning From Demonstrations},
  booktitle = {Proceedings of the Thirty-Second {AAAI} Conference on Artificial Intelligence,
               (AAAI-18), New Orleans, Louisiana, USA, February
               2-7, 2018},
  pages     = {3223--3230},
  year      = {2018}
}

@inproceedings{DBLP:conf/aaai/HasseltGS16,
  author    = {Hado van Hasselt and
               Arthur Guez and
               David Silver},
  editor    = {Dale Schuurmans and
               Michael P. Wellman},
  title     = {Deep Reinforcement Learning with Double Q-Learning},
  booktitle = {Proceedings of the Thirtieth {AAAI} Conference on Artificial Intelligence,
               February 12-17, 2016, Phoenix, Arizona, {USA}},
  pages     = {2094--2100},
  publisher = {{AAAI} Press},
  year      = {2016}
 }

@inproceedings{DBLP:conf/icml/WangSHHLF16,
  author    = {Ziyu Wang and
               Tom Schaul and
               Matteo Hessel and others},
  editor    = {Maria{-}Florina Balcan and
               Kilian Q. Weinberger},
  title     = {Dueling Network Architectures for Deep Reinforcement Learning},
  booktitle = {Proceedings of the 33nd International Conference on Machine Learning,
               {ICML} 2016, New York City, NY, USA, June 19-24, 2016},
  series    = {{JMLR} Workshop and Conference Proceedings},
  volume    = {48},
  pages     = {1995--2003},
  publisher = {JMLR.org},
  year      = {2016}
}

@article{DBLP:journals/corr/abs-1912-06680,
  author    = {Christopher Berner and
               Greg Brockman and
               Brooke Chan and others},
  title     = {Dota 2 with Large Scale Deep Reinforcement Learning},
  journal   = {CoRR},
  volume    = {abs/1912.06680},
  year      = {2019}
}

@article{DBLP:journals/jmlr/LevineFDA16,
  author    = {Sergey Levine and
               Chelsea Finn and
               Trevor Darrell and others},
  title     = {End-to-End Training of Deep Visuomotor Policies},
  journal   = {J. Mach. Learn. Res.},
  volume    = {17},
  pages     = {39:1--39:40},
  year      = {2016}
}

@article{DBLP:journals/corr/abs-1810-12894,
  author    = {Yuri Burda and
               Harrison Edwards and
               Amos J. Storkey and
               Oleg Klimov},
  title     = {Exploration by Random Network Distillation},
  journal   = {CoRR},
  volume    = {abs/1810.12894},
  year      = {2018}
}

@inproceedings{DBLP:conf/ijcai/AmirKKG16,
  author    = {Ofra Amir and
               Ece Kamar and
               Andrey Kolobov and
               Barbara J. Grosz},
  editor    = {Subbarao Kambhampati},
  title     = {Interactive Teaching Strategies for Agent Training},
  booktitle = {Proceedings of the Twenty-Fifth International Joint Conference on
               Artificial Intelligence, {IJCAI} 2016, New York, NY, USA, 9-15 July
               2016},
  pages     = {804--811},
  publisher = {{IJCAI/AAAI} Press},
  year      = {2016}
}

@inproceedings{DBLP:conf/nips/Schaal96,
  author    = {Stefan Schaal},
  editor    = {Michael Mozer and
               Michael I. Jordan and
               Thomas Petsche},
  title     = {Learning from Demonstration},
  booktitle = {Advances in Neural Information Processing Systems 9, NIPS, Denver,
               CO, USA, December 2-5, 1996},
  pages     = {1040--1046},
  publisher = {{MIT} Press},
  year      = {1996}
}

@article{DBLP:journals/corr/abs-1903-03216,
  author    = {Dong{-}Ki Kim and
               Miao Liu and
               Shayegan Omidshafiei and others},
  title     = {Learning Hierarchical Teaching in Cooperative Multiagent Reinforcement
               Learning},
  journal   = {CoRR},
  volume    = {abs/1903.03216},
  year      = {2019}
}

@inproceedings{DBLP:conf/aaai/OmidshafieiKLTR19,
  author    = {Shayegan Omidshafiei and
               Dong{-}Ki Kim and
               Miao Liu and others},
  title     = {Learning to Teach in Cooperative Multiagent Reinforcement Learning},
  booktitle = {The Thirty-Third {AAAI} Conference on Artificial Intelligence, {AAAI}
               2019, Honolulu, Hawaii,
               USA, January 27 - February 1, 2019},
  pages     = {6128--6136},
  year      = {2019}
}

@article{DBLP:journals/make/FachantidisTV19,
  author    = {Anestis Fachantidis and
               Matthew E. Taylor and
               Ioannis P. Vlahavas},
  title     = {Learning to Teach Reinforcement Learning Agents},
  journal   = {Machine Learning and Knowledge Extraction},
  volume    = {1},
  number    = {1},
  pages     = {21--42},
  year      = {2019}
}

@article{DBLP:journals/corr/abs-1712-01815,
  author    = {David Silver and
               Thomas Hubert and
               Julian Schrittwieser and others},
  title     = {Mastering Chess and Shogi by Self-Play with a General Reinforcement
               Learning Algorithm},
  journal   = {CoRR},
  volume    = {abs/1712.01815},
  year      = {2017}
}

@article{DBLP:journals/corr/abs-1903-03176,
  author    = {Kenny Young and
               Tian Tian},
  title     = {MinAtar: An Atari-inspired Testbed for More Efficient Reinforcement
               Learning Experiments},
  journal   = {CoRR},
  volume    = {abs/1903.03176},
  year      = {2019}
}

@inproceedings{DBLP:conf/iclr/FortunatoAPMHOG18,
  author    = {Meire Fortunato and
               Mohammad Gheshlaghi Azar and
               Bilal Piot and others},
  title     = {Noisy Networks For Exploration},
  booktitle = {6th International Conference on Learning Representations, {ICLR} 2018,
               Vancouver, BC, Canada, April 30 - May 3, 2018, Conference Track Proceedings},
  year      = {2018}
}

@article{DBLP:journals/corr/MnihKSGAWR13,
  author    = {Volodymyr Mnih and
               Koray Kavukcuoglu and
               David Silver and others},
  title     = {Playing Atari with Deep Reinforcement Learning},
  journal   = {CoRR},
  volume    = {abs/1312.5602},
  year      = {2013}
}

@inproceedings{DBLP:journals/corr/SchaulQAS15,
  author    = {Tom Schaul and
               John Quan and
               Ioannis Antonoglou and
               David Silver},
  title     = {Prioritized Experience Replay},
  booktitle = {4th International Conference on Learning Representations, {ICLR} 2016,
               San Juan, Puerto Rico, May 2-4, 2016, Conference Track Proceedings},
  year      = {2016}
}

@inproceedings{DBLP:conf/aaai/HesselMHSODHPAS18,
  author    = {Matteo Hessel and
               Joseph Modayil and
               Hado van Hasselt and others},
  title     = {Rainbow: Combining Improvements in Deep Reinforcement Learning},
%  booktitle = {Proceedings of the Thirty-Second {AAAI} Conference on Artificial Intelligence, (AAAI-18), New Orleans, Louisiana, USA, February 2-7, 2018},
  booktitle = {Proceedings of the Thirty-Second {AAAI} Conference on Artificial Intelligence, (AAAI-18), February 2-7, 2018},
  pages     = {3215--3222},
  year      = {2018}
}

@inproceedings{DBLP:conf/atal/SilvaGC17,
  author    = {Felipe Leno da Silva and
               Ruben Glatt and
               Anna Helena Reali Costa},
  title     = {Simultaneously Learning and Advising in Multiagent Reinforcement Learning},
%  booktitle = {Proceedings of the 16th Conference on Autonomous Agents and {Multi-Agent} Systems, {AAMAS} 2017, S{\~{a}}o Paulo, Brazil, May 8-12, 2017},
  booktitle = {Proceedings of the 16th Conference on Autonomous Agents and {Multi-Agent}
               Systems, {AAMAS} 2017, May 8-12, 2017},
  pages     = {1100--1108},
  publisher = {{ACM}},
  year      = {2017}
}

@inproceedings{aawithimitation,
  author    = {Erc{\"{u}}ment Ilhan and
               Jeremy Gow and
               Diego Perez-Liebana},
  title     = {Action Advising with Advice Imitation in Deep Reinforcement Learning},
  booktitle = {Proceedings of the 20th Conference on Autonomous Agents and {Multi-Agent}
               Systems, {AAMAS} 2021, May 3-7, 2021},
  %pages     = {1100--1108},
  publisher = {{IFAAMAS}},
  year      = {2021},
  note      = {(To appear)}
}

@inproceedings{zimmer2014teacher,
  title={Teacher-student framework: a reinforcement learning approach},
  author={Zimmer, Matthieu and Viappiani, Paolo and Weng, Paul},
  year={2014}
}

@inproceedings{DBLP:conf/atal/TorreyT13,
  author    = {Lisa Torrey and
               Matthew E. Taylor},
  title     = {Teaching on a budget: agents advising agents in reinforcement learning},
  booktitle = {International conference on Autonomous Agents and Multi-Agent Systems,
               {AAMAS} '13, Saint Paul, MN, USA, May 6-10, 2013},
  pages     = {1053--1060},
  year      = {2013}
}

@inproceedings{DBLP:conf/cig/IlhanGP19,
  author    = {Erc{\"{u}}ment Ilhan and
               Jeremy Gow and
               Diego P{\'{e}}rez{-}Li{\'{e}}bana},
  title     = {Teaching on a Budget in Multi-Agent Deep Reinforcement Learning},
  booktitle = {{IEEE} Conference on Games, CoG 2019, London, United Kingdom, August
               20-23, 2019},
  pages     = {1--8},
  year      = {2019}
}

@inproceedings{DBLP:conf/ijcai/ZhanBT16,
  author    = {Yusen Zhan and
               Haitham Bou{-}Ammar and
               Matthew E. Taylor},
  title     = {Theoretically-Grounded Policy Advice from Multiple Teachers in Reinforcement
               Learning Settings with Applications to Negative Transfer},
  booktitle = {Proceedings of the Twenty-Fifth International Joint Conference on
               Artificial Intelligence, {IJCAI} 2016, New York, NY, USA, 9-15 July
               2016},
  pages     = {2315--2321},
  year      = {2016}
}

@inproceedings{DBLP:conf/aaai/SilvaHKT20a,
  author    = {Felipe Leno da Silva and
               Pablo Hernandez{-}Leal and
               Bilal Kartal and
               Matthew E. Taylor},
  title     = {Uncertainty-Aware Action Advising for Deep Reinforcement Learning
               Agents},
  booktitle = {The Thirty-Fourth {AAAI} Conference on Artificial Intelligence, {AAAI}
               2020, The Thirty-Second Innovative Applications of Artificial Intelligence
               Conference, {IAAI} 2020, The Tenth {AAAI} Symposium on Educational
               Advances in Artificial Intelligence, {EAAI} 2020, New York, NY, USA,
               February 7-12, 2020},
  pages     = {5792--5799},
  publisher = {{AAAI} Press},
  year      = {2020}
}

@article{DBLP:journals/corr/abs-1910-12154,
  author    = {Daniel Seita and
               David M. Chan and
               Roshan Rao and others},
  title     = {{ZPD} Teaching Strategies for Deep Reinforcement Learning from Demonstrations},
  journal   = {CoRR},
  volume    = {abs/1910.12154},
  year      = {2019}
}

@inproceedings{DBLP:conf/icml/FujimotoMP19,
  author    = {Scott Fujimoto and
               David Meger and
               Doina Precup},
  title     = {Off-Policy Deep Reinforcement Learning without Exploration},
  booktitle = {Proceedings of the 36th International Conference on Machine Learning,
               {ICML} 2019, 9-15 June 2019, Long Beach, California, {USA}},
  series    = {Proceedings of Machine Learning Research},
  volume    = {97},
  pages     = {2052--2062},
  publisher = {{PMLR}},
  year      = {2019}
}

@inproceedings{DBLP:conf/nips/KumarFSTL19,
  author    = {Aviral Kumar and
               Justin Fu and others},
  title     = {Stabilizing Off-Policy Q-Learning via Bootstrapping Error Reduction},
  booktitle = {Advances in Neural Information Processing Systems 32: Annual Conference
               on Neural Information Processing Systems 2019, NeurIPS 2019, December
               8-14, 2019, Vancouver, BC, Canada},
  pages     = {11761--11771},
  year      = {2019}
}

\end{document}